\documentclass[runningheads]{llncs}
\usepackage{cite}
\usepackage{amsmath,amssymb,amsfonts}
\usepackage{booktabs}
\usepackage{colortbl}
\usepackage{graphicx}
\usepackage[breaklinks=true,colorlinks=true,linkcolor=linkcolor,citecolor=citecolor]{hyperref}
\usepackage{subfig}
\usepackage{xcolor}
\definecolor{citecolor}{HTML}{0071BC}
\definecolor{linkcolor}{HTML}{ED1C24}

\newcommand{\Eref}[1]{Equation~\eqref{#1}}

\newcommand{\Fref}[1]{Figure~\ref{#1}}

\newcommand{\Tref}[1]{Table~\ref{#1}}
\newcommand{\best}[1]{\textbf{#1}}
\newcommand{\second}[1]{\underline{#1}}

\begin{document}
\title{Adapting Vision-Language Models to Open Classes via Test-Time Prompt Tuning}
\titlerunning{Adapting VLMs to Open Classes via Test-Time Prompt Tuning}
%
\author{Zhengqing Gao\inst{1,2} \and Xiang Ao\inst{1,2} \and Xu-Yao Zhang\inst{1,2}\thanks{Corresponding author.} \and Cheng-Lin Liu\inst{1,2}}
%
%
\institute{MAIS, Institute of Automation, Chinese Academy of Sciences \and
School of Artificial Intelligence, University of Chinese Academy of Sciences
\email{\{gaozhengqing2021, aoxiang2017\}@ia.ac.cn, \{xyz, liucl\}@nlpr.ia.ac.cn}}
\maketitle              
\begin{abstract}
Adapting pre-trained models to open classes is a challenging problem in machine learning. Vision-language models fully explore the knowledge of text modality, demonstrating strong zero-shot recognition performance, which is naturally suited for various open-set problems. More recently, some research focuses on fine-tuning such models to downstream tasks. Prompt tuning methods achieved huge improvements by learning context vectors on few-shot data. However, through the evaluation under open-set adaptation setting with the test data including new classes, we find that there exists a dilemma that learned prompts have worse generalization abilities than hand-crafted prompts. In this paper, we consider combining the advantages of both and come up with a test-time prompt tuning approach, which leverages the maximum concept matching (MCM) scores as dynamic weights to generate an input-conditioned prompt for each image during test. Through extensive experiments on 11 different datasets, we show that our proposed method outperforms all comparison methods on average considering both base and new classes. The code is available at \url{https://github.com/gaozhengqing/TTPT}.

\end{abstract}
\section{Introduction}
Deep neural networks exhibit excellent capabilities when the classes in the training and test set keep the same. However, models deployed in real-world environments inevitably encounter a wide variety of out-of-distribution (OOD) data. This raises the question of how to generalize pre-trained models to open classes. Many efforts~\cite{radford2021learning,shu2023clipood} have been devoted to solve such problem, but it still remains an important challenge for developing a trustworthy machine learning system.

Recently, vision-language models have developed rapidly and achieved impressive performance on zero-shot classification, showing great potential for learning open-world visual concepts. In contrast to traditional visual models that take discrete vectors as labels in supervised learning, vision-language models use language as supervised signals yet. In this manner, the model can take full advantage of the knowledge from language modality, not limiting to closed-set concepts. Take CLIP~\cite{radford2021learning} as an example, it utilizes a contrastive objective to achieve a modality-aligned embedding space where the features of paired images and texts are pulled closer and the features of unpaired images and texts are pushed away. Then, CLIP can perform zero-shot inference by matching the embeddings of test images and textual descriptions originated from specific templates, \emph{e.g.}, replacing the [CLASS] token in ``a photo of a [CLASS]'' with candidate category names.

In order to obtain better performance on downstream tasks, the community has started to investigate how to adapt these vision-language models in a parameter-efficient way. Context Optimization (CoOp)~\cite{zhou2022learning} applies the concept of prompt learning in NLP to the fine-tuning of vision-language models, achieving considerable performance gains over prompt engineering on a wide range of datasets with only a small amount of labeled images. Specifically, CoOp converts context tokens in a prompt into several learnable vectors and optimizes them end-to-end. Despite CoOp's great success, it still leaves a problem: CoOp is apt to overfitting during training, resulting in its learned context not generalizing well. Conditional Context Optimization (CoCoOp)~\cite{zhou2022conditional} solves this problem by learning a lightweight neural network to generate a input-dependent meta token for each image, which is then added to the learnable vectors.

Motivated by the promising transfer learning capabilities, in this work, we first evaluate the performance of above methods under open-set adaptation setting where the test data contains open classes. In more detail, we fine-tune the model on base classes and test the model on all classes. It can be observed from the results that: (1) There is a huge gap between the CoOp's accuracy on base classes and new classes. (2) CoCoOp improves the new accuracy at the cost of losing part of the base accuracy. We manage to strike a better balance between the base and new accuracy by fully exploiting the information from the test data. It is widely argued that learned prompts contain more task-specific knowledge while hand-crafted prompts involve more general knowledge. From the perspective of OOD detection, we revisit this problem as distinguishing between base and new classes. With this in mind, we then propose to use the maximum concept matching (MCM) scores~\cite{ming2022delving} to measure how likely a sample is from base or new classes, and use these probabilities as weights to dynamically combine learned prompts and hand-crafted prompts. Subsequently, the combined prompts are used for inference.

In summary, our contributions are as follows:
\begin{itemize}
    \item In order to address the dilemma that arises in open-set adaptation setting where the test data includes new classes, we propose a simple but effective test-time prompt tuning method, which integrates diverse aspects of knowledge from learned and hand-crafted prompts.
    \item Our approach adopts the MCM scores to generate input-dependent weights, which are used to fuse learned and hand-crafted prompts dynamically.
    \item Extensive experiments demonstrate that our proposed method outperforms compared methods in terms of the harmonic metric. Ablation studies also validate the effectiveness of dynamic weighting and prompt fusion.
\end{itemize}

\section{Related works}
\paragraph{Vision-language models} Recently, vision-language models have received increasing attention for their ability to learn generic visual representations. Vision-language models establish connections between images and texts through a shared embedding space.

From the perspective of model architecture, vision-language models can be divided into two types. The first type of vision-language models~\cite{jia2021scaling,radford2021learning} employ two encoders to encode image and text information separately, and perform cross-modality interaction after obtaining image and text features. The selection of backbones for this type of vision-language models is more flexible, with the vision backbone usually being a ResNet~\cite{he2016deep} or ViT~\cite{dosovitskiy2021image}, and the language backbone usually being a Transformer~\cite{vaswani2017attention}. The second type of vision-language models~\cite{kim2021vilt,li2021align} adopt a deep fusion encoder to encode both image and text information and interact with each other through cross-modal attention. A Transformer backbone is often used by such vision-language models.

As for the pre-training objectives, the advances in vision-language models can be roughly grouped into two categories, namely, contrastive objectives and generative objectives. Contrastive objectives~\cite{jia2021scaling,radford2021learning} train vision-language models to learn discriminative image and text features by pulling paired images and texts close while pushing others faraway. Generative objectives learn common semantic space by generate images or texts in a manner of masked image modelling~\cite{he2022masked,bao2022beit}, masked language modelling~\cite{kenton2019bert}, and masked cross-modal modelling~\cite{singh2022flava}.

In this paper, we mainly focus on vision-language models that are pre-trained by contrastive loss functions. These models explore abundant data available from the web, \emph{e.g.}, CLIP~\cite{radford2021learning} uses 400 million accurately labeled image-text pairs and ALIGN~\cite{jia2021scaling} uses 1.8 billion noisy image-text pairs.

\paragraph{Parameter-efficient fine-tuning of vision-language models} Pre-trained vision-language models exhibit strong generalization capabilities, and fine-tuning them on downstream datasets often leads to significant performance gains compared to out-of-the-box use, \emph{i.e.}, zero-shot inference. However, due to the limited number of samples in downstream datasets, it becomes infeasible to fully fine-tune these vision-language models. Parameter-efficient fine-tuning approaches for vision-language models have been extensively studied recently. Most existing studies can be classified into two groups: prompt tuning and adapter tuning.

Prompt tuning, first developed in the NLP domain~\cite{liu2023pre}, aims to derive information from large language models that contributes to downstream tasks by learning context embeddings from finite labeled data. This technology has been borrowed for fine-tuning vision-language models on account of its parameter-efficient nature. The first way of prompt tuning is to replace the manually designed text prompts with a set of learnable text prompts and then optimize them with a few labeled downstream samples per class, \emph{e.g.} CoOp~\cite{zhou2022learning}. Prompts learned through CoOp are prone to overfitting base classes, and CoCoOp~\cite{zhou2022conditional} solves this problem by designing a meta-network to generate image-dependent prompts. The second way of prompt tuning is to add learnable visual prompts to input images. Visual prompts~\cite{bahng2022exploring} are some perturbations in the pixel space of input images, serving as cues for model decisions. The third way of prompt tuning is to modulate both image and text inputs simultaneously, and optimize them jointly in a multi-modal manner. MaPLe~\cite{khattak2023maple} introduces hierarchical prompts for both vision and language branches, where vision prompts are mapped from language prompts through a coupling function to increase the synergy between two modalities.

Instead of optimizing prompts, adapter tuning takes a different perspective by directly tuning the feature embedding through an additional lightweight adapter. CLIP-Adapter~\cite{gao2023clip} appends some bottleneck linear layers after CLIP's image and text encoders as feature adapters, and learns these adapters during few-shot fine-tuning while keeping the original model frozen. In addition, to prevent overfitting, residual connections are employed to dynamically fuse the features of the fine-tuned model and the original model. On the ground, Tip-Adapter~\cite{zhang2022tip} proposes a training-free adapter, which takes the few-shot visual features extracted by the CLIP vision encoder and the ground-truth labels as the weights of the first and second layers, respectively.

Our work focuses on how to adapt vision-language models to open classes using prompt tuning.

\paragraph{Out-of-distribution (OOD) detection of vision-language models} OOD detection is critical for models deployed in real-world environments and has been extensively studied in the fields of computer vision~\cite{hendrycks2017baseline,hendrycks2019deep,liu2020energy,hendrycks2022scaling} and natural language processing~\cite{hendrycks2020pretrained,liu2020simple} over the past few decades. For vision-language models, while earlier works~\cite{fort2021exploring,esmaeilpour2022zero} required candidate OOD labels, MCM~\cite{ming2022delving} introduces a post-hoc method that detects OOD samples by measuring the similarities between visual features and textual concepts, demonstrating promising results. More recently, some research pays more attention to few-shot OOD detection. Parameter-efficient fine-tuning methods have been validated to enhance OOD detection performance by selecting appropriate score functions~\cite{ming2023does}. LoCoOp~\cite{miyai2023locoop} proposes a training method base on prompt tuning to improve the performance of OOD detection. Notably, our method is orthogonal to OOD detection approaches and can be benefit from their advancements.

\section{Method}
\subsection{Problem Setup}
We explore the problem of adapting vision-language models to open classes. Let $\mathcal{D}_\text{train}=\{(\mathbf{x},y)\}$ denote the training data of the downstream classification task, with class label $y\in\mathcal{Y}=\{1,2,\cdots,K\}$, and $\mathcal{D}_\text{test}=\mathcal{D}_\text{base}\cup\mathcal{D}_\text{new}$ denote the test data, which contains new classes that do not appear in the training data. Here, $\mathcal{D}_\text{base}=\{(\mathbf{x},y)\}$ represents the test data from base classes, with class label $y\in\mathcal{Y}=\{1,2,\cdots,K\}$, and $\mathcal{D}_\text{new}=\{(\mathbf{x}^\prime,y^\prime)\}$ represents the test data from new classes, with class label $y^\prime\in\mathcal{Y}^\prime=\{K+1,K+2,\cdots,K^\prime\}$. Given a pre-trained vision-language model, it is first fine-tuned on $\mathcal{D}_\text{train}$ in a few-shot fashion. The fine-tuned model is expected to achieve not only good generalization on $\mathcal{D}_\text{base}$ but also good adaptation on $\mathcal{D}_\text{new}$.

\subsection{Preliminaries}
Recently, vision-language models have shown tremendous potential for a variety of computer vision tasks. In this paper, we employ the pioneering work CLIP~\cite{radford2021learning} as a representative. CLIP adopts a dual-stream architecture, which consists of an image encoder $g_I(\cdot)$ and a text encoder $g_T(\cdot)$ to encode image and text modalities, respectively, to a common feature space. During the pre-training phase, CLIP learns transferable feature representations from a large-scale image caption dataset via a multi-modal contrastive loss that promotes the alignment of features from different modalities.

CLIP shows promising zero-shot generalization performance on various downstream tasks. Specifically, given an input image $\mathbf{x}$, let $\{\mathbf{t}_i\}_{i=1}^K$ be a set of prompts, each representing a category, where $K$ is the number of categories. Notably, each $\mathbf{t}_i$ is generated from a prompt (\emph{e.g.}, ``a photo of a [CLASS]'') by replacing the [CLASS] token with the corresponding category name. Then, the probability that $\mathbf{x}$ belongs to category $k$ can be calculated as
\begin{equation}
    p(y=k\mid\mathbf{x})=\frac{\exp(\text{sim}(g_I(\mathbf{x}),g_T(\mathbf{t}_k))/\tau)}{\sum_{i=1}^K\exp(\text{sim}(g_I(\mathbf{x}),g_T(\mathbf{t}_i))/\tau)},
\end{equation}
where $\text{sim}(\cdot,\cdot)$ represents the cosine similarity between embeddings, and $\tau$ represents the temperature parameter learned by CLIP.

In order to improve the performance of CLIP on downstream tasks, various parameter-efficient fine-tuning methods have been proposed. Among them, prompt tuning demonstrates performance improvements compared to zero-shot settings. While CLIP uses hand-designed prompt templates, prompt tuning optimizes the word embeddings of the prompts. Specifically, a typical method CoOp introduces $M$ learnable context vectors $\{\mathbf{v}_i\}_{i=1}^M$, each having the same dimension with the word embeddings. The prompt for class $k$ is obtained by concatenating the context vectors and the word embedding $\mathbf{w}_k$ for the classname, \emph{i.e.}, $\mathbf{t}_k=[\mathbf{v}_1,\mathbf{v}_2,\dots,\mathbf{v}_M,\mathbf{w}_k]$. All classes share the context vectors. To optimize the context vectors, a cross-entropy loss is used as the learning objective. The image encoder and the text encoder are frozen during the training process.

CLIP not only demonstrates remarkable zero-shot classification abilities but also exhibits excellent performance in zero-shot out-of-distribution (OOD) detection. It is note worth noting that here in-distribution (ID) classes are defined as those used in the fine-tuning phase rather than the pre-training phase, and OOD classes are defined accordingly based on the ID classes. MCM~\cite{ming2022delving} explore the OOD detection capabilities of the pre-trained CLIP without fine-tuning on the ID dataset. Concretely, the MCM score is written as
\begin{equation}
    S_\text{MCM}(\mathbf{x})=\max_{k=1}^K\frac{\exp(\text{sim}(g_I(\mathbf{x}),g_T(\mathbf{t}_k))/\tau)}{\sum_{i=1}^K\exp(\text{sim}(g_I(\mathbf{x}),g_T(\mathbf{t}_i))/\tau)},
\end{equation}
where temperature $\tau$ is a hyperparameter that needs to be tuned on the downstream dataset. Because of the CLIP's well-aligned multi-modal feature space, ID images will match one of the candidate texts with a high score, and vice versa. Formally, OOD detection can be achieved by:
\begin{equation}
    f_\lambda(\mathbf{x})=\begin{cases}
        \text{ID}&S_\text{MCM}(\mathbf{x})\geq\lambda\\
        \text{OOD}&S_\text{MCM}(\mathbf{x})<\lambda
    \end{cases},
\end{equation}
where $\lambda$ is a threshold that can be chosen as needed in practice, \emph{e.g.}, a large percentage of ID data is above the threshold.

\begin{figure}[t]
    \centering
    \subfloat{\includegraphics[width=0.9\textwidth]{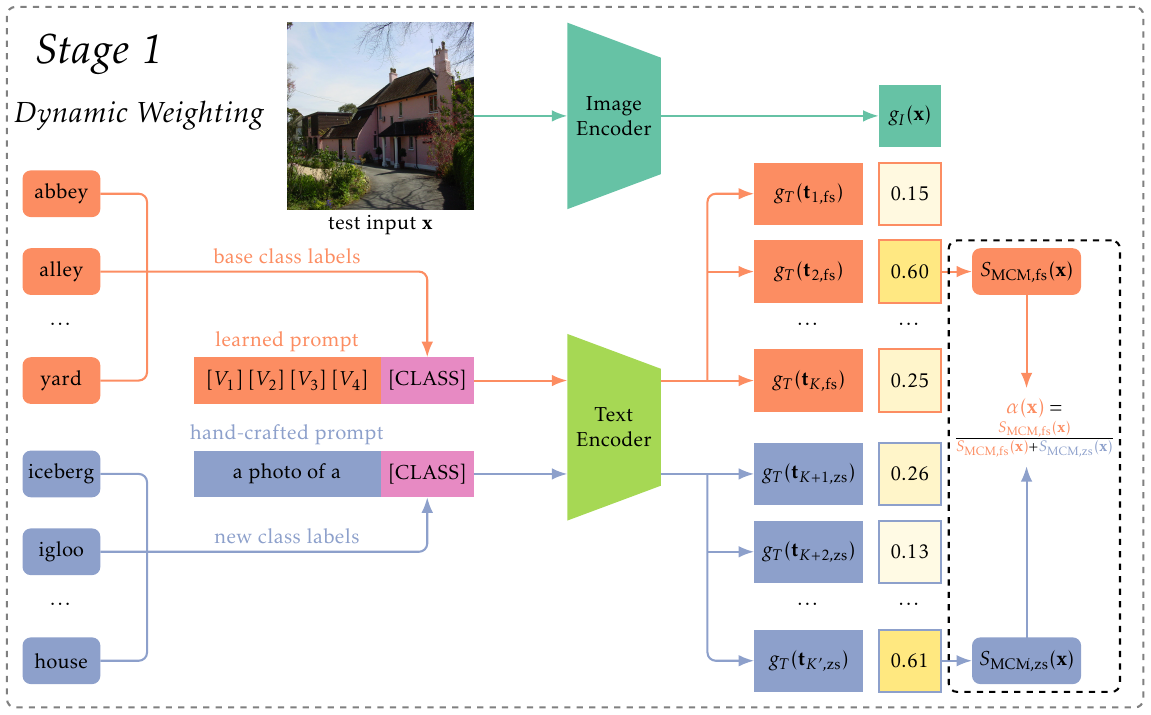}}
    \\
    \subfloat{\includegraphics[width=0.9\textwidth]{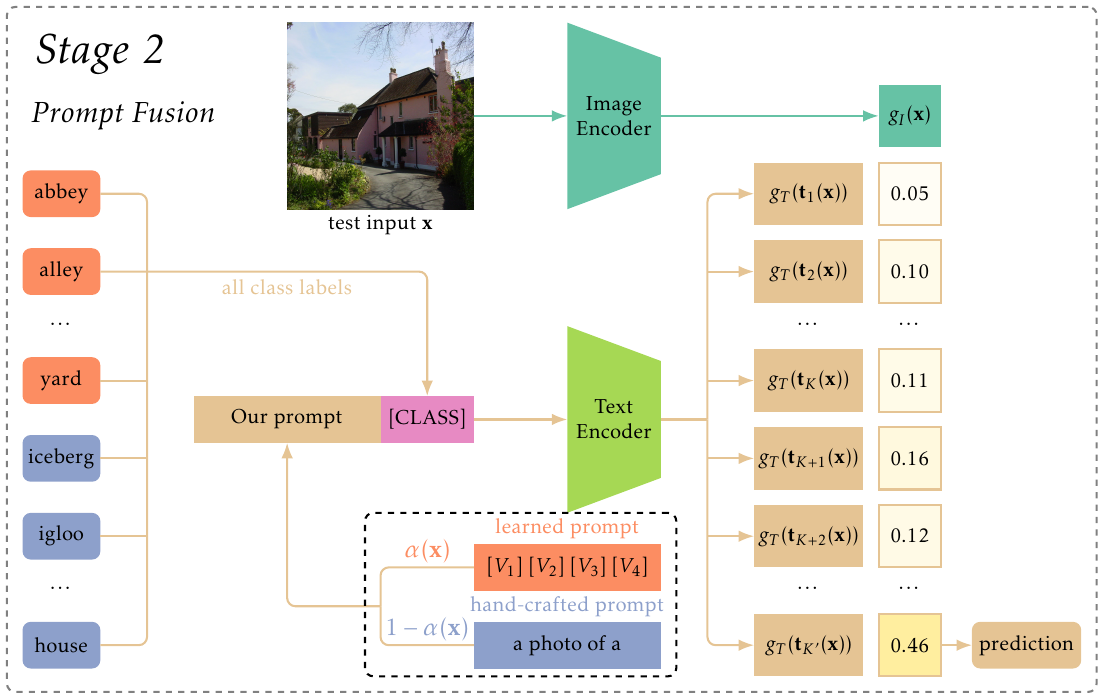}}
    \caption{Overview of our proposed method. Given a test input $\mathbf{x}$, we first compute the MCM scores using learned prompt and hand-crafted prompt on base and new classes, respectively. Then, a dynamic weighting strategy is employed to generate the input-dependent prompt that combines the advantages of learned prompt and hand-crafted prompt. It is worth noting that the hand-crafted prompt is dataset-specific, which is ``a photo of a [CLASS]." for ImageNet while ``a photo of a [CLASS], a type of flower." for Flowers102, for example. The learned prompt has the same dimension and length with the corresponding hand-crafted prompt.}
    \label{fig:overview}
\end{figure}

\subsection{Proposed Approach}
While prompt tuning leads to better performance on downstream tasks, it impairs the generalization ability of CLIP. That is to say, prompt tuning performs well on base classes, but poorly on new classes. We argue that hand-designed contexts have better generalization ability because they avoid overfitting to a specific set of classes during training. In this paper, we aim to improve the accuracy of CLIP when the test data contains open classes by utilizing both learned contexts and hand-designed contexts. We propose a test-time prompt tuning method, and an overview is shown in \Fref{fig:overview}.

In the first stage, to measure the likelihood that test images are from base and new classes, we calculate their MCM scores on base and new classes, respectively. As pointed out in a recent research~\cite{ming2023does} that prompt tuning based-methods can improve OOD detection performance after few-shot fine-tuning, we employ learned prompts to get the MCM scores of input images on base classes. Contrastively, we obtain the MCM scores of input images on new classes via hand-crafted prompts due to their superior generalization capabilities. Formally, given an input image $\mathbf{x}$, let $\mathbf{t}_{i,\text{zs}}$ be the hand-crafted prompt of class $i$ for zero-shot inference, and $\mathbf{t}_{i,\text{fs}}$ be the learned prompt of class $i$ through few-shot prompt tuning. Both $\mathbf{t}_{i,\text{zs}}$ and $\mathbf{t}_{i,\text{fs}}$ share the same dimension and length. The MCM score on base classes $S_\text{MCM,fs}(\mathbf{x})$ and new classes $S_\text{MCM,zs}(\mathbf{x})$ are calculated as
\begin{align}
    S_\text{MCM,fs}(\mathbf{x})&=\max_{k=1}^K\frac{\exp(\text{sim}(g_I(\mathbf{x}),g_T(\mathbf{t}_{k,\text{fs}}))/\tau)}{\sum_{i=1}^{K}\exp(\text{sim}(g_I(\mathbf{x}),g_T(\mathbf{t}_{i,\text{fs}}))/\tau)},\label{eq:MCM_fs}\\
    S_\text{MCM,zs}(\mathbf{x})&=\max_{k=K+1}^{K^\prime}\frac{\exp(\text{sim}(g_I(\mathbf{x}),g_T(\mathbf{t}_{k,\text{zs}}))/\tau)}{\sum_{i=K+1}^{K^\prime}\exp(\text{sim}(g_I(\mathbf{x}),g_T(\mathbf{t}_{i,\text{zs}}))/\tau)}.\label{eq:MCM_zs}
\end{align}

In the second stage, we design a dynamic weighting strategy to generate a specific prompt for each image. More concretely, the MCM scores are treated as sample-level weights for fusing the learned prompts with the hand-designed prompts. The prompt for class $i$ is thus input-dependent:
\begin{equation}
    \mathbf{t}_i(\mathbf{x})=\alpha(\mathbf{x})\mathbf{t}_{i,\text{fs}}+(1-\alpha(\mathbf{x}))\mathbf{t}_{i,\text{zs}},
\end{equation}
where
\begin{equation}
    \alpha(\mathbf{x})=\frac{S_{\text{MCM,fs}}(\mathbf{x})}{S_{\text{MCM,fs}}(\mathbf{x})+S_{\text{MCM,zs}}(\mathbf{x})}.
\end{equation}
Then, the prediction probability is calculated on all classes:
\begin{equation}
    p(y=k\mid\mathbf{x})=\frac{\exp(\text{sim}(g_I(\mathbf{x}),g_T(\mathbf{t}_k(\mathbf{x})))/\tau)}{\sum_{i=1}^{K^\prime}\exp(\text{sim}(g_I(\mathbf{x}),g_T(\mathbf{t}_i(\mathbf{x})))/\tau)}.
\end{equation}

\section{Experiments}
\subsection{Setup}
\paragraph{Datasets} Following~\cite{zhou2022learning,zhou2022conditional}, We conduct experiments on 11 diverse image recognition datasets, which cover different vision tasks. In particular, the benchmark provides ImageNet~\cite{deng2009imagenet} and Caltech101~\cite{fei2004learning} for generic object classification; OxfordPets~\cite{parkhi2012cats}, StanfordCars~\cite{krause20133d}, Flowers102~\cite{nilsback2008automated}, Food101~\cite{bossard2014food} and FGVCAircraft~\cite{maji2013fine} for fine-grained classification; SUN397~\cite{xiao2010sun} for scene recognition; DTD~\cite{cimpoi2014describing} for texture classification; EuroSAT~\cite{helber2019eurosat} for satellite image recognition and UCF101~\cite{soomro2012ucf101} for action recognition. For each dataset, we randomly sample a few-shot subset of the training set for fine-tuning and use the full test set for evaluation, as in~\cite{zhou2022conditional}.

\paragraph{Training details} The implementation of our method is build upon the official code of CoOp\footnote{\url{https://github.com/KaiyangZhou/CoOp}}. We use ViT-B/16 as the vision backbone of CLIP by default, with the number of shots set to 16. We split the classes into base classes and new classes equally, and train models on the base classes only, as in~\cite{zhou2022conditional}.

The difference is that we conduct evaluation jointly on the base and new classes, while~\cite{zhou2022conditional} evaluate separately on the base and new classes. That is to say, our candidate set in the test phase contains all classes.

First, we follow~\cite{zhou2022learning} to obtain the learned prompts. Specifically, the learnable context vectors are initialized by the embeddings of the corresponding templates. We use an SGD optimizer during training. The learning rate is set to 0.02 initially and gradually decays through a cosine annealing scheduler. We set the maximum epoch to 50 for ImageNet and 200 for others, and the trained model in the last epoch is selected for evaluation. In addition, we also employ the constant warmup trick by setting the learning rate to 1e-5 in the first epoch to prevent gradient explosion. Then, we set the temperature in \Eref{eq:MCM_fs} and \Eref{eq:MCM_zs} to the default of CLIP models, \emph{i.e.}, 0.01 for CLIP-B/16.

For evaluation, we report the accuracy of the base and new classes, as well as their harmonic mean
\begin{equation}
    \text{H}=\frac{2\times\text{base}\times\text{new}}{\text{base}+\text{new}}.
\end{equation}

\paragraph{Baselines} We compare our proposed approach with seven baseline methods. \textbf{CLIP}~\cite{radford2021learning} uses hand-crafted prompts for zero-shot inference, which are carefully designed for different datasets. \textbf{CoOp}~\cite{zhou2022learning} replaces the hand-designed contexts of the CLIP text branch with a set of continuous learnable vectors, which can be optimized in a few-shot fashion. \textbf{CoCoOp}~\cite{zhou2022conditional} generates input-dependent context tokens for each image by learning a lightweight neural network, which solves the problem that CoOp is prone to overfitting during training. \textbf{SHIP}~\cite{wang2023improving} is a plug-and-play generative approach that improves new accuracy by synthesizing features of new classes using only category names. \textbf{CLIP-Adapter}~\cite{gao2023clip} employs additional bottleneck layers behind CLIP's encoders to learn new features and blend the new features with the original features via residual connection. \textbf{Tip-Adapter}~\cite{zhang2022tip} creates a key-value cache model from the few-shot training set as a two-layer adapter, which does not require any training.

\begin{table}[t]
    \centering
    \caption{Adaptation performance on 11 downstream datasets with open classes. The \textbf{bold} values indicate the best results, and the \underline{underlined} values indicate the second best results.}
    \label{tab:open classes}
    \subfloat[\textbf{Average.}]{
        \resizebox{0.3\textwidth}{!}{%
            \begin{tabular}{lccc}
                \toprule
                 & Base & New & H \\ \midrule
                CLIP & 63.4 & \best{67.2} & 65.3 \\
                CoOp & 75.9 & 55.4 & 64.1 \\
                CoCoOp & 73.8 & 63.9 & \second{68.5} \\
                CoOp + SHIP & 78.8 & 57.6 & 66.5 \\
                CLIP-Adapter & \best{82.8} & 30.5 & 44.6 \\
                CLIP-Adapter + SHIP & \second{82.5} & 35.7 & 49.9 \\
                Tip-Adapter + SHIP & 82.1 & 49.0 & 61.4 \\
                \rowcolor[HTML]{EFEFEF} 
                Ours & 75.3 & \second{64.1} & \best{69.3} \\ \bottomrule
            \end{tabular}%
        }
    }
    \subfloat[ImageNet.]{
        \resizebox{0.3\textwidth}{!}{%
            \begin{tabular}{lccc}
                \toprule
                 & Base & New & H \\ \midrule
                CLIP & 68.4 & 65.1 & 66.7 \\
                CoOp & 72.9 & 64.8 & 68.6 \\
                CoCoOp & 72.6 & \best{67.4} & \second{69.9} \\
                CoOp + SHIP & 72.1 & 64.3 & 68.0 \\
                CLIP-Adapter & 71.9 & 65.0 & 68.3 \\
                CLIP-Adapter + SHIP & 72.0 & 66.2 & 69.0 \\
                Tip-Adapter + SHIP & \best{75.5} & 60.8 & 67.3 \\
                \rowcolor[HTML]{EFEFEF} 
                Ours & \second{73.0} & \second{67.2} & \best{70.0} \\ \bottomrule
            \end{tabular}%
        }
    }
    \subfloat[Caltech101.]{
        \resizebox{0.3\textwidth}{!}{%
            \begin{tabular}{lccc}
                \toprule
                 & Base & New & H \\ \midrule
                CLIP & 94.0 & \best{92.1} & 93.0 \\
                CoOp & 95.5 & 80.0 & 87.1 \\
                CoCoOp & 95.5 & 91.2 & \second{93.3} \\
                CoOp + SHIP & 97.6 & 88.2 & 92.6 \\
                CLIP-Adapter & \second{98.1} & 71.1 & 82.4 \\
                CLIP-Adapter + SHIP & 97.6 & 77.2 & 86.2 \\
                Tip-Adapter + SHIP & \best{98.3} & 81.6 & 89.1 \\
                \rowcolor[HTML]{EFEFEF} 
                Ours & 95.5 & \second{91.7} & \best{93.6} \\ \bottomrule
            \end{tabular}%
        }
    }
    \\
    \subfloat[OxfordPets.]{
        \resizebox{0.3\textwidth}{!}{%
            \begin{tabular}{lccc}
                \toprule
                 & Base & New & H \\ \midrule
                CLIP & 84.8 & \best{93.7} & 89.0 \\
                CoOp & 88.4 & 87.5 & 87.9 \\
                CoCoOp & 91.7 & \second{92.7} & \best{92.2} \\
                CoOp + SHIP & \best{94.1} & 83.8 & 88.6 \\
                CLIP-Adapter & 91.7 & 33.3 & 48.9 \\
                CLIP-Adapter + SHIP & 91.8 & 40.8 & 56.5 \\
                Tip-Adapter + SHIP & \second{94.0} & 83.2 & 88.3 \\
                \rowcolor[HTML]{EFEFEF} 
                Ours & 92.0 & 91.9 & \second{91.9} \\ \bottomrule
            \end{tabular}%
        }
    }
    \subfloat[StanfordCars.]{
        \resizebox{0.3\textwidth}{!}{%
            \begin{tabular}{lccc}
                \toprule
                 & Base & New & H \\ \midrule
                CLIP & 60.0 & \best{71.1} & 65.1 \\
                CoOp & \second{73.2} & 57.7 & 64.5 \\
                CoCoOp & 67.5 & 65.3 & 66.4 \\
                CoOp + SHIP & 69.2 & 65.2 & \second{67.1} \\
                CLIP-Adapter & 79.3 & 34.4 & 47.9 \\
                CLIP-Adapter + SHIP & 78.5 & 39.1 & 52.2 \\
                Tip-Adapter + SHIP & \best{79.8} & 51.8 & 62.9 \\
                \rowcolor[HTML]{EFEFEF} 
                Ours & 70.7 & \second{66.8} & \best{68.7} \\ \bottomrule
            \end{tabular}%
        }
    }
    \subfloat[Flowers102.]{
        \resizebox{0.3\textwidth}{!}{%
            \begin{tabular}{lccc}
                \toprule
                 & Base & New & H \\ \midrule
                CLIP & 67.1 & \best{73.4} & 70.1 \\
                CoOp & 97.0 & 52.9 & 68.5 \\
                CoCoOp & 88.0 & 64.3 & 74.3 \\
                CoOp + SHIP & 93.5 & 62.4 & \second{74.9} \\
                CLIP-Adapter & \best{98.4} & 27.7 & 43.2 \\
                CLIP-Adapter + SHIP & \second{97.7} & 34.3 & 50.7 \\
                Tip-Adapter + SHIP & 95.4 & 35.3 & 51.5 \\
                \rowcolor[HTML]{EFEFEF} 
                Ours & 95.1 & \second{67.0} & \best{78.6} \\ \bottomrule
            \end{tabular}%
        }
    }
    \\
    \subfloat[Food101.]{
        \resizebox{0.3\textwidth}{!}{%
            \begin{tabular}{lccc}
                \toprule
                 & Base & New & H \\ \midrule
                CLIP & 85.9 & \second{85.9} & \second{85.9} \\
                CoOp & 82.4 & 76.9 & 79.6 \\
                CoCoOp & 86.5 & \best{87.1} & \best{86.8} \\
                CoOp + SHIP & 88.3 & 81.9 & 84.9 \\
                CLIP-Adapter & \second{88.4} & 44.5 & 59.2 \\
                CLIP-Adapter + SHIP & 88.3 & 51.7 & 65.2 \\
                Tip-Adapter + SHIP & \best{89.8} & 75.7 & 82.2 \\
                \rowcolor[HTML]{EFEFEF} 
                Ours & 84.9 & 85.7 & 85.3 \\ \bottomrule
            \end{tabular}%
        }
    }
    \subfloat[FGVCAircraft.]{
        \resizebox{0.3\textwidth}{!}{%
            \begin{tabular}{lccc}
                \toprule
                 & Base & New & H \\ \midrule
                CLIP & 20.0 & \best{29.7} & 23.9 \\
                CoOp & 31.3 & 20.9 & 25.1 \\
                CoCoOp & 27.6 & \second{26.1} & \best{26.8} \\
                CoOp + SHIP & 33.2 & 13.8 & 19.5 \\
                CLIP-Adapter & \best{42.5} & 8.6 & 14.3 \\
                CLIP-Adapter + SHIP & \second{42.2} & 10.3 & 16.5 \\
                Tip-Adapter + SHIP & 41.5 & 14.6 & 21.6 \\
                \rowcolor[HTML]{EFEFEF} 
                Ours & 27.7 & 24.4 & \second{25.9} \\ \bottomrule
            \end{tabular}%
        }
    }
    \subfloat[SUN397.]{
        \resizebox{0.3\textwidth}{!}{%
            \begin{tabular}{lccc}
                \toprule
                 & Base & New & H \\ \midrule
                CLIP & 60.2 & \second{65.0} & 62.5 \\
                CoOp & 72.0 & 51.1 & 59.8 \\
                CoCoOp & 70.6 & \best{67.2} & \best{68.9} \\
                CoOp + SHIP & 75.7 & 60.2 & \second{67.1} \\
                CLIP-Adapter & \best{79.1} & 20.4 & 32.4 \\
                CLIP-Adapter + SHIP & \second{79.0} & 27.6 & 40.9 \\
                Tip-Adapter + SHIP & 76.6 & 51.8 & 61.8 \\
                \rowcolor[HTML]{EFEFEF} 
                Ours & 71.9 & 60.2 & 65.5 \\ \bottomrule
            \end{tabular}%
        }
    }
    \\
    \subfloat[DTD.]{
        \resizebox{0.3\textwidth}{!}{%
            \begin{tabular}{lccc}
                \toprule
                 & Base & New & H \\ \midrule
                CLIP & 42.0 & \best{46.4} & 44.1 \\
                CoOp & 66.1 & 33.1 & 44.1 \\
                CoCoOp & 61.7 & \second{43.2} & \best{50.8} \\
                CoOp + SHIP & 74.1 & 32.1 & 44.8 \\
                CLIP-Adapter & \second{81.9} & 2.9 & 5.6 \\
                CLIP-Adapter + SHIP & \best{82.3} & 7.9 & 14.3 \\
                Tip-Adapter + SHIP & 80.0 & 15.8 & 26.4 \\
                \rowcolor[HTML]{EFEFEF} 
                Ours & 61.5 & 40.9 & \second{49.1} \\ \bottomrule
            \end{tabular}%
        }
    }
    \subfloat[EuroSAT.]{
        \resizebox{0.3\textwidth}{!}{%
            \begin{tabular}{lccc}
                \toprule
                 & Base & New & H \\ \midrule
                CLIP & 51.2 & \best{45.8} & 48.3 \\
                CoOp & 75.7 & 38.9 & \second{51.4} \\
                CoCoOp & 74.9 & 33.4 & 46.2 \\
                CoOp + SHIP & 86.3 & 18.9 & 31.0 \\
                CLIP-Adapter & \best{93.6} & 0.1 & 0.1 \\
                CLIP-Adapter + SHIP & \second{93.1} & 0.4 & 0.9 \\
                Tip-Adapter + SHIP & 89.8 & 10.4 & 18.7 \\
                \rowcolor[HTML]{EFEFEF} 
                Ours & 79.5 & \second{43.2} & \best{56.0} \\ \bottomrule
            \end{tabular}%
        }
    }
    \subfloat[UCF101.]{
        \resizebox{0.3\textwidth}{!}{%
            \begin{tabular}{lccc}
                \toprule
                 & Base & New & H \\ \midrule
                CLIP & 64.2 & \best{71.0} & 67.4 \\
                CoOp & 80.1 & 46.0 & 58.4 \\
                CoCoOp & 75.2 & 64.9 & 69.7 \\
                CoOp + SHIP & 82.6 & 62.6 & \best{71.2} \\
                CLIP-Adapter & \best{85.8} & 27.5 & 41.6 \\
                CLIP-Adapter + SHIP & \second{85.4} & 37.6 & 52.2 \\
                Tip-Adapter + SHIP & 82.1 & 58.3 & 68.2 \\
                \rowcolor[HTML]{EFEFEF} 
                Ours & 76.6 & \second{66.5} & \second{71.2} \\ \bottomrule
            \end{tabular}%
        }
    }
    \vspace{-1em}
\end{table}

\begin{figure}[t]
    \centering
    \subfloat[]{\label{fig:coop}\includegraphics[width=0.5\textwidth]{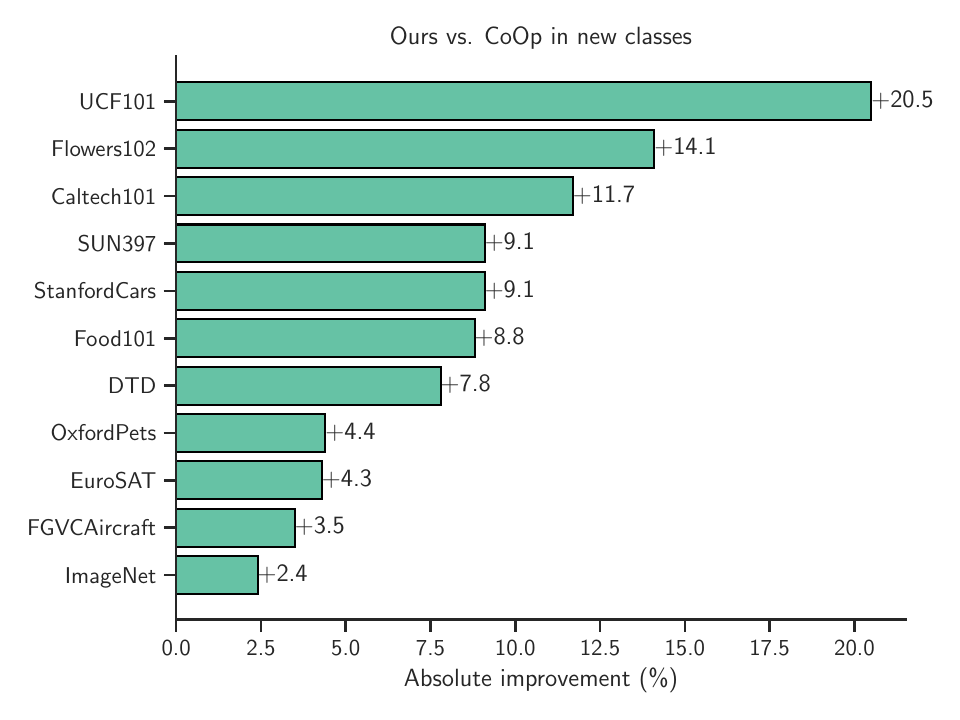}}
    \subfloat[]{\label{fig:cocoop}\includegraphics[width=0.5\textwidth]{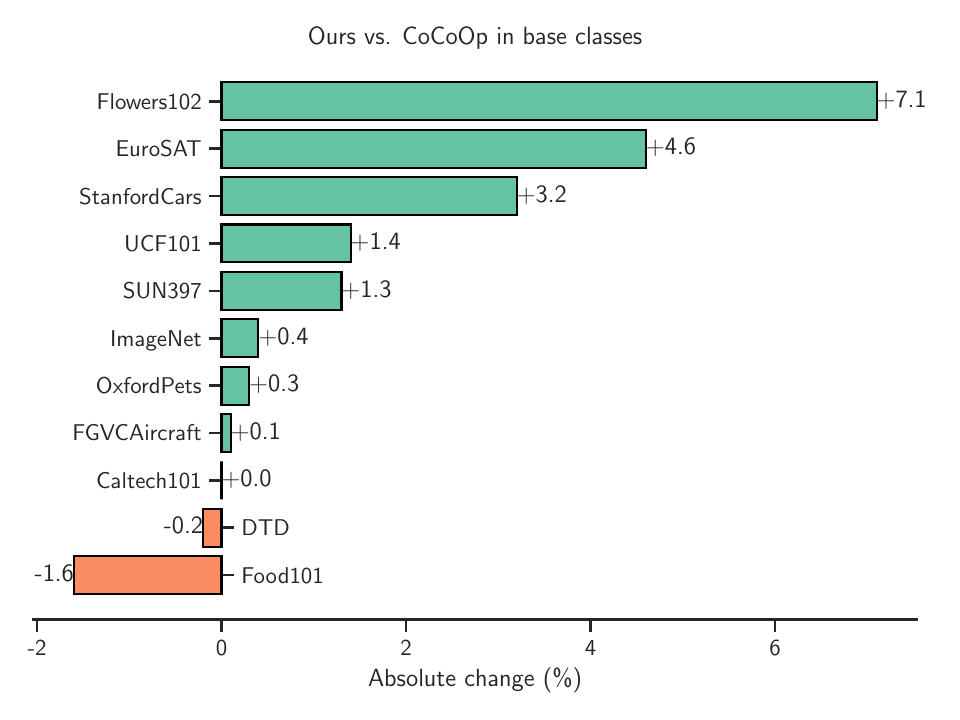}}
    \caption{Comparisons of our method with CoOp and CoCoOp.}
    \label{fig:comparisons}
\end{figure}

\subsection{Comparison with state-of-the-art methods}
We evaluate the performance of our method with seven baseline methods when adapting to open classes, and the detailed results are presented in \Tref{tab:open classes}.

Despite CoOp's excellent performance on base classes, its accuracy on new classes has declined dramatically, resulting in even worse overall performance than CLIP. This can be attributed to the fact that CoOp is overfitted to a limited number of training samples from base classes. Our approach, however, alleviates this problem significantly, as shown in \Fref{fig:coop}. For example, compared to CLIP, CoOp's average new accuracy over 11 datasets drops by 11.8\%, while our method only decreased by 3.1\%. CoCoOp improves the accuracy of new classes using more generalizable instance-conditioned prompts at the expense of a partial decrease in base accuracy. From \Fref{fig:cocoop} we can observe that our approach further maintains better performance on base classes. Numerically, the average base accuracy of CoCoOp on 11 datasets is 1.9\% lower than that of CoOp, whereas our method is merely 0.4\% lower.

Compared to prompt tuning methods, adapter tuning methods show a much larger performance degradation in this more realistic setting. Take CLIP-Adapter as an example, it has an average new accuracy of only 30.5\% on 11 datasets, and even more exaggerated, it has an average new accuracy of 0.1\% on EuroSAT dataset. Applying SHIP to above methods can alleviate the problem to some extent, while our method can improve the performance even more.

\begin{figure}[t]
    \centering
    \includegraphics[width=0.5\textwidth]{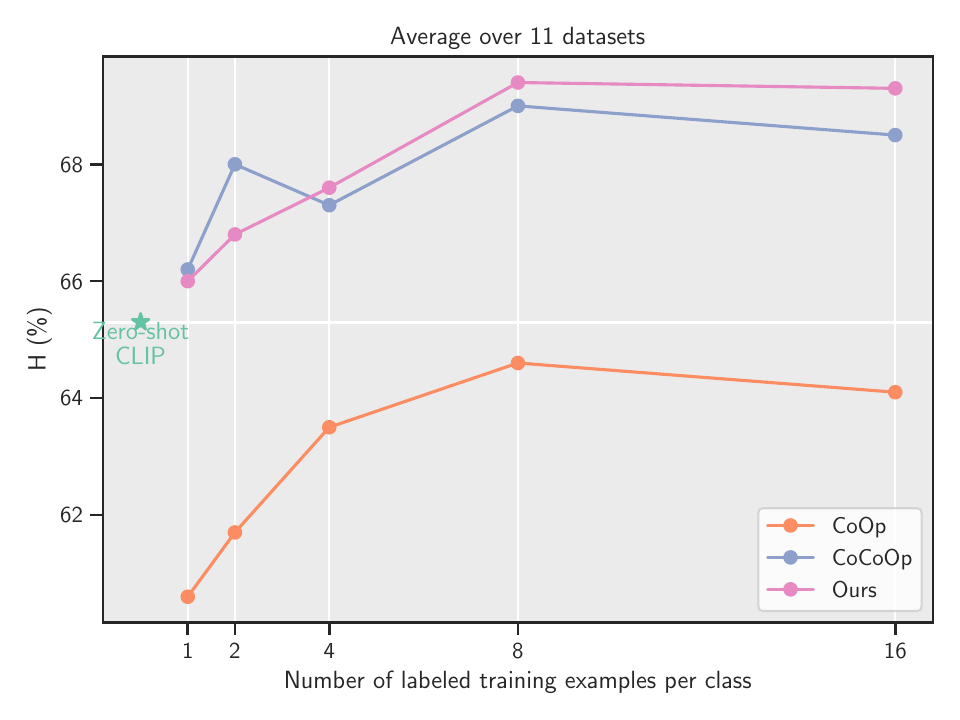}
    \caption{Performance of different methods with varying number of shots.}
    \label{fig:shots}
\end{figure}

\begin{figure}[t]
    \centering
    \includegraphics[width=0.5\textwidth]{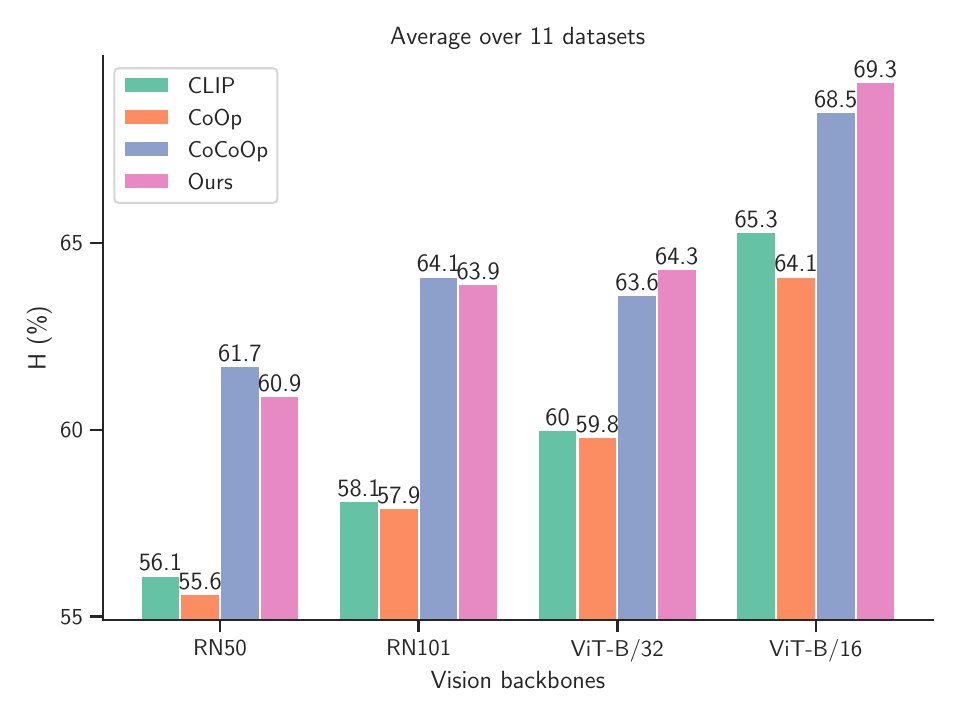}
    \caption{Performance of different methods with diverse vision backbones.}
    \label{fig:backbones}
\end{figure}

\begin{table}[t]
    \centering
    \caption{Ablation studies. We evaluate the effectiveness of dynamic weighting and prompt fusion. Average accuracy over 11 datasets is reported.}
    \label{tab:ablation}
    \begin{tabular}{lccc}
        \toprule
         & Base & New & H \\ \midrule
        Input-independent prompt fusion & 73.6 & 64.0 & 68.5 \\
        Classifier combinations & 68.5 & 66.5 & 67.5 \\
        Ours & 75.3 & 64.1 & 69.3 \\ \bottomrule
    \end{tabular}%
\end{table}

\begin{table}[t]
    \centering
    \caption{Adaptation performance with different temperatures. Average accuracy over 11 datasets is reported.}
    \label{tab:temperature}
    \begin{tabular}{lccc}
        \toprule
         & Base & New & H \\ \midrule
        1 & 73.8 & 63.9 & 68.5 \\
        0.1 & 74.9 & 63.5 & 68.7 \\
        0.01 (Ours) & 75.3 & 64.1 & 69.3 \\ \bottomrule
    \end{tabular}%
\end{table}

\subsection{Further Analysis}

\paragraph{Comparison with input-independent prompt fusion} First, we think of input-independent prompt fusion, specifically, $\alpha(\mathbf{x})$ is set to 1/2. It can be concluded from \Tref{tab:ablation} that it is superior to use the MCM scores as weights.

\paragraph{Comparison with classifier combinations} Second, we consider combining classifiers in Stage 1, that is, the text features of base and new classes are obtained by learned and hand-crafted prompts, respectively, and then the similarities between the image feature and text features of all classes are calculated. The results in \Tref{tab:ablation} demonstrate the effectiveness of our proposed dynamic weighting strategy.

\paragraph{Number of shots} We explore the performance of different methods under few-shot evaluation protocol with various number of shots. We set the number of shots to 1, 2, 4, 8 and 16, respectively. The experimental results are summarized in \Fref{fig:shots}. We observe increased performance when the number of shots changes from 1 to 8, while leveled or even decreased performance when the number of shots changes from 8 to 16. This can be ascribed to the fact that as the number of shots grows, the overfitting becomes more severe, causing a rise in base accuracy and a decline in new accuracy.

\paragraph{Vision backbones} A CLIP model can use either a CNN~\cite{he2016deep} or a ViT~\cite{dosovitskiy2021image} as its vision backbone, and \Fref{fig:backbones} shows the impact of different vision backbones. We can make two observations. First, all methods exhibit enhanced performance as the vision backbone advances, and the gaps between our method and comparison methods increase. Second, CoCoOp performs better on CNNs, while our method performs better on ViTs.

\paragraph{Effects of temperature} We explore the impact of temperature in \Eref{eq:MCM_fs} and \Eref{eq:MCM_zs}, and the results are summarized in \Tref{tab:temperature}. As noted in MCM~\cite{ming2022delving}, choosing a moderate temperature for softmax scaling on cosine similarity scores improves the ID-OOD separability. Specifically, we vary the temperature from 0.01 to 1. We observe optimal performance at a temperature of 0.01.

\section{Conclusion}
With the development of vision-language models, our research timely addresses an important problem of adapting these models to open classes. Prompt tuning approaches tend to overfit the training data, leading to poor generalization. We propose a dynamic weighting scheme which integrates the task-specific knowledge of learned prompts with the general knowledge of hand-crafted prompts, improving performance effectively. We hope our work can inspire and motivate further studies on the adaptation of vision-language models.

\paragraph{Acknowledgements} This work has been supported by the National Science and Technology Major Project (2022ZD0116500), National Natural Science Foundation of China (U20A20223, 62222609, 62076236), CAS Project for Young Scientists in Basic Research (YSBR-083), and Key Research Program of Frontier Sciences of CAS (ZDBS-LY-7004).
%
%
%
\bibliographystyle{splncs04}
\bibliography{mybibliography}

\begin{thebibliography}{10}
\providecommand{\url}[1]{\texttt{#1}}
\providecommand{\urlprefix}{URL }
\providecommand{\doi}[1]{https://doi.org/#1}

\bibitem{bahng2022exploring}
Bahng, H., Jahanian, A., Sankaranarayanan, S., Isola, P.: Exploring visual
  prompts for adapting large-scale models. arXiv preprint arXiv:2203.17274
  (2022)

\bibitem{bao2022beit}
Bao, H., Dong, L., Piao, S., Wei, F.: Beit: Bert pre-training of image
  transformers. In: ICLR (2022)

\bibitem{bossard2014food}
Bossard, L., Guillaumin, M., Van~Gool, L.: Food-101--mining discriminative
  components with random forests. In: ECCV (2014)

\bibitem{cimpoi2014describing}
Cimpoi, M., Maji, S., Kokkinos, I., Mohamed, S., Vedaldi, A.: Describing
  textures in the wild. In: CVPR (2014)

\bibitem{deng2009imagenet}
Deng, J., Dong, W., Socher, R., Li, L.J., Li, K., Fei-Fei, L.: Imagenet: A
  large-scale hierarchical image database. In: CVPR (2009)

\bibitem{dosovitskiy2021image}
Dosovitskiy, A., Beyer, L., Kolesnikov, A., Weissenborn, D., Zhai, X.,
  Unterthiner, T., Dehghani, M., Minderer, M., Heigold, G., Gelly, S., et~al.:
  An image is worth 16x16 words: Transformers for image recognition at scale.
  In: ICLR (2021)

\bibitem{esmaeilpour2022zero}
Esmaeilpour, S., Liu, B., Robertson, E., Shu, L.: Zero-shot out-of-distribution
  detection based on the pre-trained model clip. In: AAAI (2022)

\bibitem{fei2004learning}
Fei-Fei, L., Fergus, R., Perona, P.: Learning generative visual models from few
  training examples: An incremental bayesian approach tested on 101 object
  categories. In: CVPRW (2004)

\bibitem{fort2021exploring}
Fort, S., Ren, J., Lakshminarayanan, B.: Exploring the limits of
  out-of-distribution detection. In: NeurIPS (2021)

\bibitem{gao2023clip}
Gao, P., Geng, S., Zhang, R., Ma, T., Fang, R., Zhang, Y., Li, H., Qiao, Y.:
  Clip-adapter: Better vision-language models with feature adapters. IJCV
  (2023)

\bibitem{he2022masked}
He, K., Chen, X., Xie, S., Li, Y., Doll{\'a}r, P., Girshick, R.: Masked
  autoencoders are scalable vision learners. In: CVPR (2022)

\bibitem{he2016deep}
He, K., Zhang, X., Ren, S., Sun, J.: Deep residual learning for image
  recognition. In: CVPR (2016)

\bibitem{helber2019eurosat}
Helber, P., Bischke, B., Dengel, A., Borth, D.: Eurosat: A novel dataset and
  deep learning benchmark for land use and land cover classification. JSTARS
  (2019)

\bibitem{hendrycks2022scaling}
Hendrycks, D., Basart, S., Mazeika, M., Zou, A., Kwon, J., Mostajabi, M.,
  Steinhardt, J., Song, D.: Scaling out-of-distribution detection for
  real-world settings. In: ICML (2022)

\bibitem{hendrycks2017baseline}
Hendrycks, D., Gimpel, K.: A baseline for detecting misclassified and
  out-of-distribution examples in neural networks. In: ICLR (2017)

\bibitem{hendrycks2020pretrained}
Hendrycks, D., Liu, X., Wallace, E., Dziedzic, A., Krishnan, R., Song, D.:
  Pretrained transformers improve out-of-distribution robustness. In: ACL
  (2020)

\bibitem{hendrycks2019deep}
Hendrycks, D., Mazeika, M., Dietterich, T.: Deep anomaly detection with outlier
  exposure. In: ICLR (2019)

\bibitem{jia2021scaling}
Jia, C., Yang, Y., Xia, Y., Chen, Y.T., Parekh, Z., Pham, H., Le, Q., Sung,
  Y.H., Li, Z., Duerig, T.: Scaling up visual and vision-language
  representation learning with noisy text supervision. In: ICML (2021)

\bibitem{kenton2019bert}
Kenton, J.D.M.W.C., Toutanova, L.K.: Bert: Pre-training of deep bidirectional
  transformers for language understanding. In: NAACL (2019)

\bibitem{khattak2023maple}
Khattak, M.U., Rasheed, H., Maaz, M., Khan, S., Khan, F.S.: Maple: Multi-modal
  prompt learning. In: CVPR (2023)

\bibitem{kim2021vilt}
Kim, W., Son, B., Kim, I.: Vilt: Vision-and-language transformer without
  convolution or region supervision. In: ICML (2021)

\bibitem{krause20133d}
Krause, J., Stark, M., Deng, J., Fei-Fei, L.: 3d object representations for
  fine-grained categorization. In: ICCVW (2013)

\bibitem{li2021align}
Li, J., Selvaraju, R., Gotmare, A., Joty, S., Xiong, C., Hoi, S.C.H.: Align
  before fuse: Vision and language representation learning with momentum
  distillation. In: NeurIPS (2021)

\bibitem{liu2020simple}
Liu, J., Lin, Z., Padhy, S., Tran, D., Bedrax~Weiss, T., Lakshminarayanan, B.:
  Simple and principled uncertainty estimation with deterministic deep learning
  via distance awareness. In: NeurIPS (2020)

\bibitem{liu2023pre}
Liu, P., Yuan, W., Fu, J., Jiang, Z., Hayashi, H., Neubig, G.: Pre-train,
  prompt, and predict: A systematic survey of prompting methods in natural
  language processing. ACM Computing Surveys  (2023)

\bibitem{liu2020energy}
Liu, W., Wang, X., Owens, J., Li, Y.: Energy-based out-of-distribution
  detection. In: NeurIPS (2020)

\bibitem{maji2013fine}
Maji, S., Rahtu, E., Kannala, J., Blaschko, M., Vedaldi, A.: Fine-grained
  visual classification of aircraft. arXiv preprint arXiv:1306.5151  (2013)

\bibitem{ming2022delving}
Ming, Y., Cai, Z., Gu, J., Sun, Y., Li, W., Li, Y.: Delving into
  out-of-distribution detection with vision-language representations. In:
  NeurIPS (2022)

\bibitem{ming2023does}
Ming, Y., Li, Y.: How does fine-tuning impact out-of-distribution detection for
  vision-language models? IJCV  (2023)

\bibitem{miyai2023locoop}
Miyai, A., Yu, Q., Irie, G., Aizawa, K.: Locoop: Few-shot out-of-distribution
  detection via prompt learning. In: NeurIPS (2023)

\bibitem{nilsback2008automated}
Nilsback, M.E., Zisserman, A.: Automated flower classification over a large
  number of classes. In: ICVGIP (2008)

\bibitem{parkhi2012cats}
Parkhi, O.M., Vedaldi, A., Zisserman, A., Jawahar, C.: Cats and dogs. In: CVPR
  (2012)

\bibitem{radford2021learning}
Radford, A., Kim, J.W., Hallacy, C., Ramesh, A., Goh, G., Agarwal, S., Sastry,
  G., Askell, A., Mishkin, P., Clark, J., et~al.: Learning transferable visual
  models from natural language supervision. In: ICML (2021)

\bibitem{shu2023clipood}
Shu, Y., Guo, X., Wu, J., Wang, X., Wang, J., Long, M.: Clipood: Generalizing
  clip to out-of-distributions. In: ICML (2023)

\bibitem{singh2022flava}
Singh, A., Hu, R., Goswami, V., Couairon, G., Galuba, W., Rohrbach, M., Kiela,
  D.: Flava: A foundational language and vision alignment model. In: CVPR
  (2022)

\bibitem{soomro2012ucf101}
Soomro, K., Zamir, A.R., Shah, M.: Ucf101: A dataset of 101 human actions
  classes from videos in the wild. arXiv preprint arXiv:1212.0402  (2012)

\bibitem{vaswani2017attention}
Vaswani, A., Shazeer, N., Parmar, N., Uszkoreit, J., Jones, L., Gomez, A.N.,
  Kaiser, {\L}., Polosukhin, I.: Attention is all you need. In: NeurIPS (2017)

\bibitem{wang2023improving}
Wang, Z., Liang, J., He, R., Xu, N., Wang, Z., Tan, T.: Improving zero-shot
  generalization for clip with synthesized prompts. In: ICCV (2023)

\bibitem{xiao2010sun}
Xiao, J., Hays, J., Ehinger, K.A., Oliva, A., Torralba, A.: Sun database:
  Large-scale scene recognition from abbey to zoo. In: CVPR (2010)

\bibitem{zhang2022tip}
Zhang, R., Fang, R., Zhang, W., Gao, P., Li, K., Dai, J., Qiao, Y., Li, H.:
  Tip-adapter: Training-free clip-adapter for better vision-language modeling.
  In: ECCV (2022)

\bibitem{zhou2022conditional}
Zhou, K., Yang, J., Loy, C.C., Liu, Z.: Conditional prompt learning for
  vision-language models. In: CVPR (2022)

\bibitem{zhou2022learning}
Zhou, K., Yang, J., Loy, C.C., Liu, Z.: Learning to prompt for vision-language
  models. IJCV  (2022)

\end{thebibliography}
%




\end{document}